\documentclass[preprint, 12pt]{elsarticle} 

\biboptions{numbers,sort&compress}
\usepackage{siunitx}
\usepackage{booktabs}
\usepackage{caption}
\usepackage{subcaption}
\usepackage{float}
\usepackage{tabularx}
\usepackage{adjustbox}
\usepackage{amsmath}
\usepackage{algorithm}
\usepackage{algpseudocode}
\usepackage[hidelinks]{hyperref}
\usepackage{xcolor}
\renewcommand{\textcolor}[2]{#2} 
\usepackage{acronym}
\usepackage{multicol}
\usepackage{array}
\usepackage{makecell}
\usepackage{amssymb}
\usepackage{graphicx}
\usepackage{xurl}
\usepackage{tikz}
\usetikzlibrary{arrows.meta,positioning,calc}

\makeatletter
\def\ps@pprintTitle{%
  \let\@oddhead\@empty
  \let\@evenhead\@empty
  \def\@oddfoot{}%
  \let\@evenfoot\@oddfoot}
\makeatother

\begin{document}

\let\WriteBookmarks\relax
\def\floatpagepagefraction{1}
\def\textpagefraction{.001}

\renewcommand{\topfraction}{0.9}
\renewcommand{\bottomfraction}{0.8}
\renewcommand{\textfraction}{0.07}
\renewcommand{\floatpagefraction}{0.7}
\setcounter{topnumber}{3}
\setcounter{bottomnumber}{2}
\setcounter{totalnumber}{4}
\sloppy

\title{Validation-Gated Multi-Agent Governance for Online Adaptation of Thermal--Hydraulic Surrogate Models under Operating-Regime Shift}

\author[1]{Doyeong Lim\corref{cor1}}
\ead{dylim24@unist.ac.kr}
\author[1]{Seungyoon Lee}
\author[1]{In Cheol Bang\corref{cor1}}
\ead{icbang@unist.ac.kr}

\affiliation[1]{
  organization={Department of Nuclear Engineering, Ulsan National Institute of Science and Technology (UNIST)},
  city={Ulsan},
  postcode={44919},
  country={Republic of Korea}
}

\cortext[cor1]{Corresponding authors}

\begin{abstract}
  Artificial-intelligence surrogates can support second-by-second thermal--hydraulic forecasting, but models selected and frozen offline may become condition-locked once deployed outside their pretraining envelope. This study develops a guarded continual-adaptation framework for experimental thermal--hydraulic loop data in which role-separated agents---Monitor, Diagnosis, Adaptation, Safety-Auditor, and Orchestrator---diagnose error signatures, prioritize candidate model families, and review promotions, while deterministic champion--challenger gates and background shadow learning retain final authority over model replacement. Seven surrogate families were screened by blocked three-fold cross-validation, and a temporal Fourier neural operator was selected as the initial champion for 60-s-history-to-10-s-trajectory forecasting on two held-out transients, with three seeds per adaptive mode. Static deployment gave a channel-averaged MAE of 7.06 and a 56.8\% warning-exceedance ratio; rule-based adaptation reduced MAE to 6.54, whereas shadow refresh alone remained close to Static. The MA-Full mode, in which the role-separated multi-agent council reviews every evaluated stream step, achieved the lowest mean error, 5.72, and 35.8\% exceedance, corresponding to a 19.0\% improvement over Static. Paired bootstrap intervals against Static excluded zero, although intervals among adaptive modes overlapped and the six paired units limit broad statistical claims. Validated promotions from the neural operator to Transformer and graph neural network indicate that logged, gate-controlled adaptation can support auditable surrogate evolution while deterministic gates retain deployment authority.

\end{abstract}



\maketitle

\section{Introduction}

Real-time prediction of loop-scale thermal--hydraulic (TH) states is of growing interest for experimental operation support, digital-twin monitoring, passive-safety assessment, and anomaly interpretation. High-fidelity system codes and computational-fluid-dynamics models remain indispensable for licensing studies and uncertainty quantification, but their computational cost makes them impractical as multi-output forecasters within a second-by-second experimental stream. Artificial-intelligence (AI) surrogates have been proposed to address this need by learning the delayed coupling among heater power, control actions, temperatures, pressures, and flow rates from prior simulation or measurement records, returning multi-step predictions at low serving latency.

Across a range of nuclear thermal--hydraulic problems, a modest accuracy penalty in exchange for low-latency inference has motivated surrogate deployment. Early studies established the speed argument, with deep networks trained on simulated loss-of-coolant accidents reproducing simulation accuracy at much lower computational cost~\citep{radaideh2020loca}. The approach was subsequently extended to severe-accident progression through rolling-window recurrent forecasters that capture mitigation-induced hysteresis and reconstruct lost signals~\citep{lee2024rolling,song2024mlsevere}, to physics-aware and operator-learning architectures that embed conservation constraints or quantify predictive uncertainty as a built-in out-of-distribution warning~\citep{antonello2023pinn,cheng2025fbdeeponet}, and to deployment-oriented digital twins in which neural-operator inference makes second-by-second twin updating computationally feasible~\citep{daniell2025digitaltwin,kobayashi2024neuraloperator,lim2025ai}. In most of these studies, however, the surrogate is trained, validated, and then \emph{frozen} offline, and is evaluated only within the same operating envelope used for training. This practice leaves open how a frozen surrogate behaves once the operating envelope shifts away from the pretraining conditions.

The relevant question in deployment is therefore not whether a surrogate interpolates well within its training distribution, but whether it remains reliable once the facility enters a regime that was poorly represented during pretraining. TH facilities are particularly susceptible to such shifts: thermal inertia, flow redistribution, passive heat-removal pathways, and control-policy changes can rapidly alter the mapping from recent measurements to future states, which can leave a pretrained surrogate \emph{condition-locked} and lead to smooth but physically inaccurate trajectories. This stability--plasticity tension has been formalized in the streaming-learning literature as concept drift and online continual learning~\citep{gama2014conceptdrift,lu2019conceptdrift,gunasekara2023oscl,bidaki2025ocl,wang2024clsurvey}, supported by drift detectors~\citep{page1954cusum,gama2004ddm,bifet2007adwin} and replay- or regularization-based update rules~\citep{kirkpatrick2017ewc,aljundi2019mir,buzzega2020der}; production-ML practice similarly reports that frozen models accumulate hidden technical debt as data drift~\citep{sculley2015mldebt,polyzotis2018datalifecycle}, and process-industry soft sensors are commonly adapted online through moving-window and just-in-time learning~\citep{zhang2023softsensor}. In nuclear TH, online calibration has been demonstrated for digital twins~\citep{song2022onlinecalibration}, but most surrogates, including recent Transformer/Informer fault-diagnosis models~\citep{wang2024hybridtransformer,wang2024informernpp}, retain a fixed offline backbone and do not specify how the model should respond when the operating envelope changes.

Continual adaptation, however, cannot be reduced to updating the surrogate whenever its error rises. Aggressive online learning can amplify sensor faults, overfit the most recent window, and induce forgetting of earlier regimes~\citep{kirkpatrick2017ewc,wang2024clsurvey}; it can also promote a challenger that is locally accurate but globally fragile, which is an undesirable failure mode in safety-relevant monitoring. A more conservative response is deterministic guarded adaptation based on thresholds, cooldowns, bounded fine-tuning, post-update validation, and rollback, following the MLOps champion--challenger pattern~\citep{sculley2015mldebt,polyzotis2018datalifecycle}. Such rules constrain when and how the model may change, but on their own they do not explain why a particular error signature has appeared, decide which model family should be added to the candidate pool, or veto a short-term improvement that conceals tail-risk behavior. Recent role-based agent and multi-agent frameworks address related decision-support tasks by decomposing monitoring, diagnosis, planning, and review into specialized reasoning--acting roles~\citep{wu2024autogen,yao2023react,shinn2023reflexion,park2023generativeagents,hong2024metagpt,gottweis2025coscientist,gridach2025agentic,guo2024llmma}, building on chain-of-thought prompting as the underlying reasoning primitive~\citep{wei2022cot}. In engineering modeling, such agents have automated end-to-end data-driven workflows---data preprocessing, network construction, training, hyperparameter search, and uncertainty quantification---through self-correcting code-repair loops, with both supervisor-orchestrated multi-agent and ReAct single-agent designs reaching performance comparable to benchmark reference solutions on a standardized thermal--hydraulic benchmark~\citep{liu2026automating}. In nuclear applications, LLM agents have been coupled with reactor digital twins to support remote monitoring and operator-facing control assistance~\citep{ndum2026large}, reflecting growing interest in incorporating language-model reasoning into safety-relevant system operation. A streaming plant surrogate, however, is not an open-ended reasoning task, and direct agent control of a safety-relevant model would not be appropriate. This raises the question of how agent recommendations can be bounded within a deterministic safety envelope, so that agents guide continual surrogate evolution without acting as the real-time controller.

This paper addresses that gap using experimental data from the TH facility. First, we propose a guarded multi-agent governance framework in which Monitor, Diagnosis, Adaptation, Safety-Auditor, and Orchestrator roles support a champion--challenger surrogate pool, while deterministic validation and rollback rules retain final authority over model replacement. Second, we construct the initial champion through a blocked 3-fold cross-validation screen over seven candidate architectures---LSTM~\citep{hochreiter1997lstm}, GRU~\citep{cho2014gru}, Transformer~\citep{vaswani2017attention}, neural ODE~\citep{chen2018node}, graph neural network~\citep{battaglia2018graph,corso2024graph}, DeepONet~\citep{lu2021deeponet}, and temporal Fourier neural operator~\citep{li2021fno}---so that the streaming baseline is a competitive offline model. Third, we conduct a seven-mode streaming comparison (Static, Rule-H, Shadow, Single-H, Single-Full, MA-H, and MA-Full) on operating transients with three seeds per mode, and quantify paired differences using 95\% bootstrap confidence intervals~\citep{efron1979bootstrap}. This design separates the contributions of rule-only safeguards, scheduled shadow refresh, single-agent planning, and role-separated multi-agent governance. The remainder of the paper is organized as follows. Section~2 describes the dataset, forecasting task, offline candidate pool, governance loop, and safety gates. Section~3 reports offline champion selection, streaming results, validated champion transitions, and limitations. Section~4 concludes.

\section{Methodology}

\begin{figure}[htbp]
  \centering
  \includegraphics[width=\linewidth]{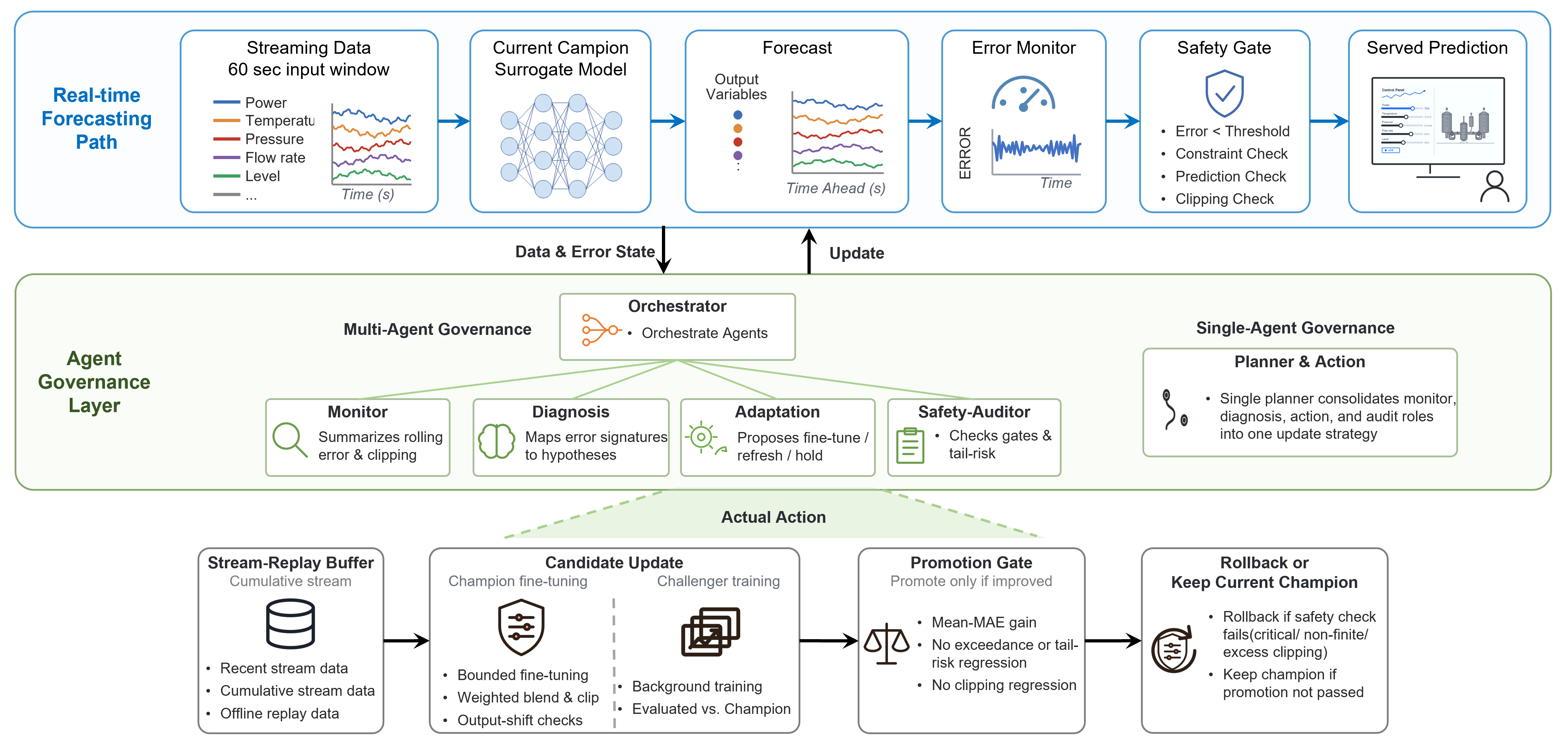}
  \caption{Multi-agent continual-adaptation framework with real-time inference.}
  \label{fig:agentarchitecture}
\end{figure}

Figure~\ref{fig:agentarchitecture} summarizes the framework, which separates a serving path from an adaptation path. In the serving path, the accepted champion receives a 60-s multivariate TH history and returns a 10-s multi-output forecast. Real-time inference continues while the monitoring layer records physical-unit error, warning state, clipping behavior, and dominant error channels. The adaptation path is separate from this serving path. Recent stream samples and cumulative replay data generate bounded fine-tuned models or background challengers, but a candidate can serve predictions only after passing validation, robustness, and promotion gates. When enabled, agents provide diagnosis, planning, veto, and audit support inside this envelope; they do not directly override the serving model.

\subsection{Experimental facility and dataset}

\begin{figure}[htbp]
  \centering
  \includegraphics[width=0.92\linewidth]{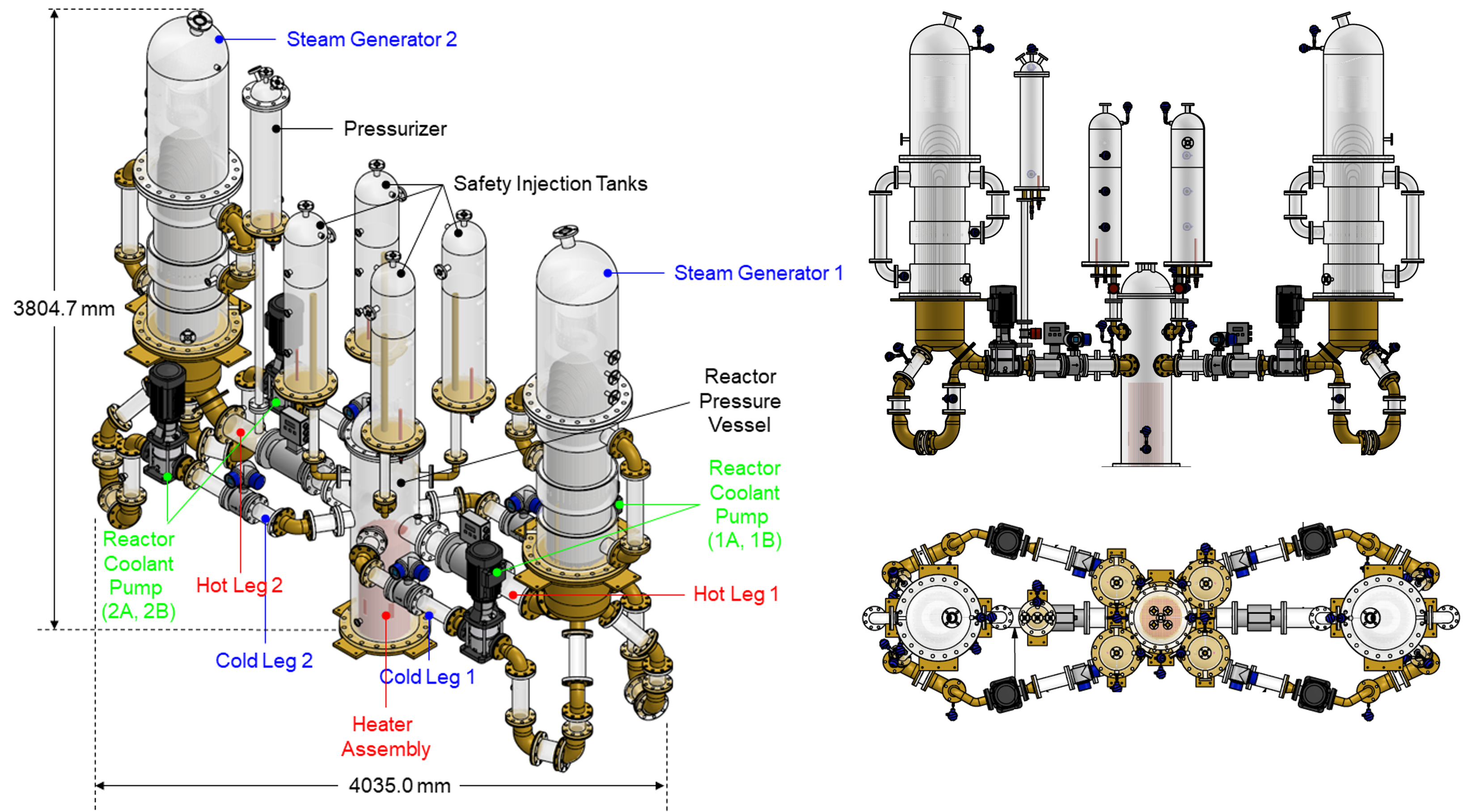}
  \caption{Thermal--hydraulic test facility used for data generation~\citep{kim2021urilo}.}
  \label{fig:urilo}
\end{figure}

The experimental data were obtained from the TH loop designed for transparent visualization, passive-safety experiments, and system-scale sensing studies~\citep{kim2021urilo}. The loop in Figure~\ref{fig:urilo} contains major pressurized-water reactor components, including the reactor pressure vessel, steam generators, hot and cold legs, pressurizer, safety injection tank, and primary-system piping. The measurements are sampled at one-second resolution and include heater power, control signals, temperatures, pressures, flow rates, and vacuum-related channels.

\begin{figure}[htbp]
  \centering
  \includegraphics[width=0.92\linewidth]{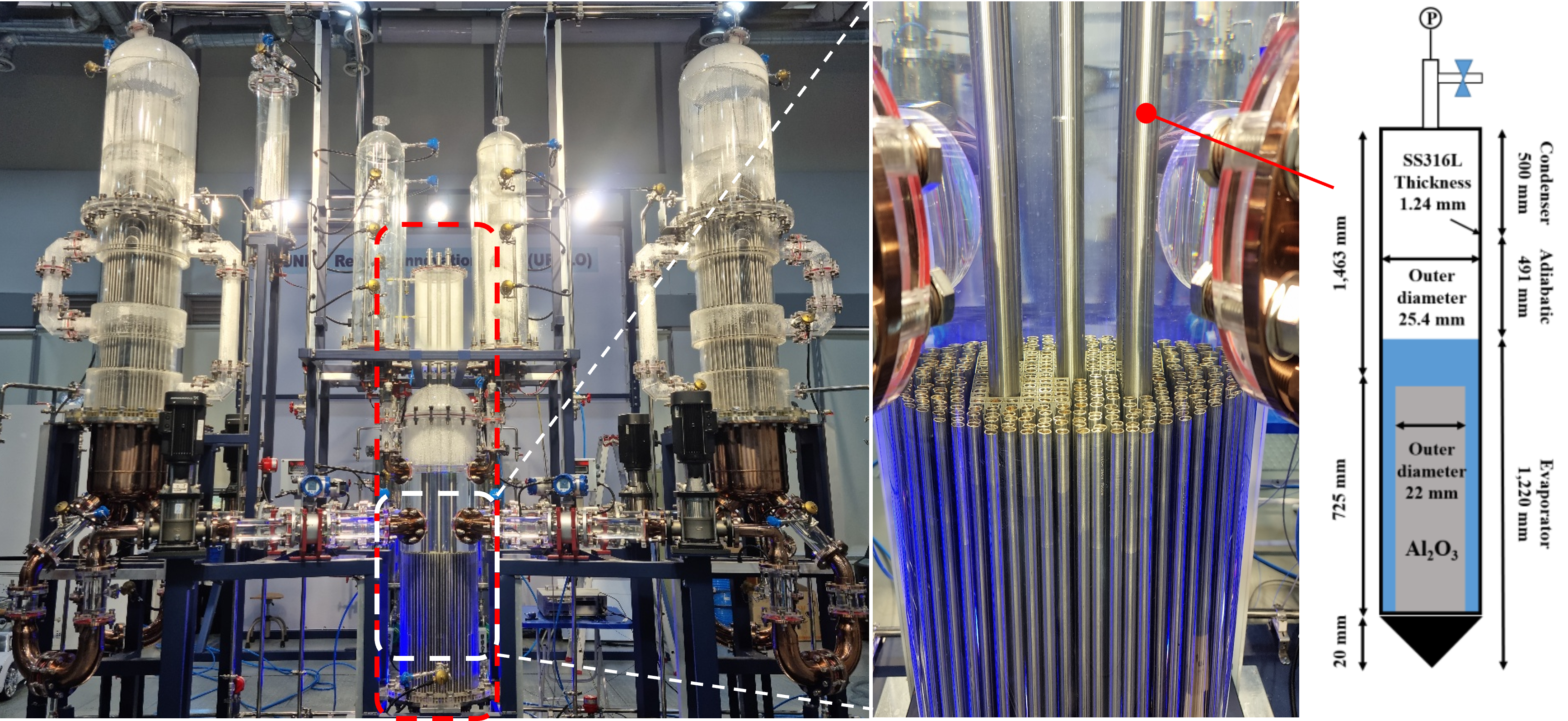}
  \caption{Heat-pipe insertion configuration used as the held-out envelope-shift test condition~\citep{kim2022urilohp}.}
  \label{fig:urilohp}
\end{figure}

The study is organized as an offline-to-online deployment experiment rather than a random train--test split. The offline stage uses normal-operation, power-reduction, and station-blackout (SBO) records (8{,}410 steps) for model-family screening, hyperparameter selection, and initial champion pretraining. The online stage then deploys the selected champion on two held-out transients outside the offline envelope. The first is a heat-pipe insertion configuration in Figure~\ref{fig:urilohp}, which adds a passive heat-removal path in the upper reactor region and alters the coupling among temperatures, pressures, flow redistribution, and heater response. The second is a separate held-out operating transient recorded on the same loop without any hardware modification; it follows a power and flow schedule that is not represented in the offline records, providing an independent out-of-envelope test that does not involve the heat-pipe change. The two splits provide 2{,}701 and 3{,}185 stream windows and are each restarted from the same pretrained champion. Table~\ref{tab:dataset_summary} summarizes the partition and settings.

\begin{table*}[htbp]
  \centering
  \caption{Experimental dataset and surrogate model training method.}
  \label{tab:dataset_summary}
  \footnotesize
  \setlength{\tabcolsep}{4pt}
  \renewcommand{\arraystretch}{1.18}
  \begin{tabular}{@{}p{0.15\linewidth}p{0.35\linewidth}p{0.13\linewidth}p{0.3\linewidth}@{}}
    \toprule
    Category                     & Condition or setting                                                                                                                                                                                         & Length                           & Use in this study                                                                 \\
    \midrule
    Offline pretraining data     & Normal operation, power-reduction operation, and SBO scenario experiments.                                                                                                                                   & 8{,}410 steps                    & Offline architecture search, pretraining, and initial champion selection.         \\[2pt]
    Held-out streaming test data & Two held-out operating transients: heat-pipe insertion in the upper reactor region, and a second transient (separate campaign, no hardware change) on an operating schedule absent from the offline records. & 2{,}701 + 3{,}185 stream windows & Streaming evaluation, guarded adaptation, and out-of-envelope robustness testing. \\[2pt]
    Pretraining setting          & 1000 epochs, early-stopping patience 100, batch size 32, blocked 3-fold cross-validation, and RTX 3090 GPU.                                                                                                  & --                               & Model-family and hyperparameter selection before online deployment.               \\[2pt]
    Forecasting task             & 60~s state-feedback input window and 10~s future trajectory prediction.                                                                                                                                      & --                               & 105 input channels and 77 valid supervised thermal--hydraulic targets.            \\
    \bottomrule
  \end{tabular}
\end{table*}

\subsection{State-feedback forecasting task and preprocessing}

The surrogate is formulated as a state-feedback multi-step forecaster rather than a purely control-conditioned predictor. For each streaming index $t$, the input is a 60~s history window
\begin{equation}
  \mathbf{X}_t = \left[\mathbf{u}_{t-59:t},\,\mathbf{z}_{t-59:t}\right] \in \mathbb{R}^{60\times105},
\end{equation}
where $\mathbf{u}$ denotes control and power-related variables and $\mathbf{z}$ denotes measured plant-state variables. The output is the next 10~s trajectory
\begin{equation}
  \widehat{\mathbf{Y}}_t = f_{\theta}(\mathbf{X}_t) \in \mathbb{R}^{10\times77},
\end{equation}
corresponding to valid thermal--hydraulic targets such as temperatures, pressures, and flow rates. Including the recent measured state allows the model to encode loop memory, delayed thermal transport, and thermal inertia, which are essential for short-horizon thermal--hydraulic forecasting.

Known unreliable sensor channels are treated conservatively: they are retained as zero-filled input positions so that the deployed interface keeps a fixed channel schema, but are excluded from the supervised target set and the training/validation loss. All reported MAE and RMSE values are computed after inverse transformation and then averaged across channels, so that threshold exceedance and rollback decisions use the same scale as online monitoring.

\subsection{Offline surrogate search and initial champion}

\begin{figure}[htbp]
  \centering
  \includegraphics[width=\linewidth]{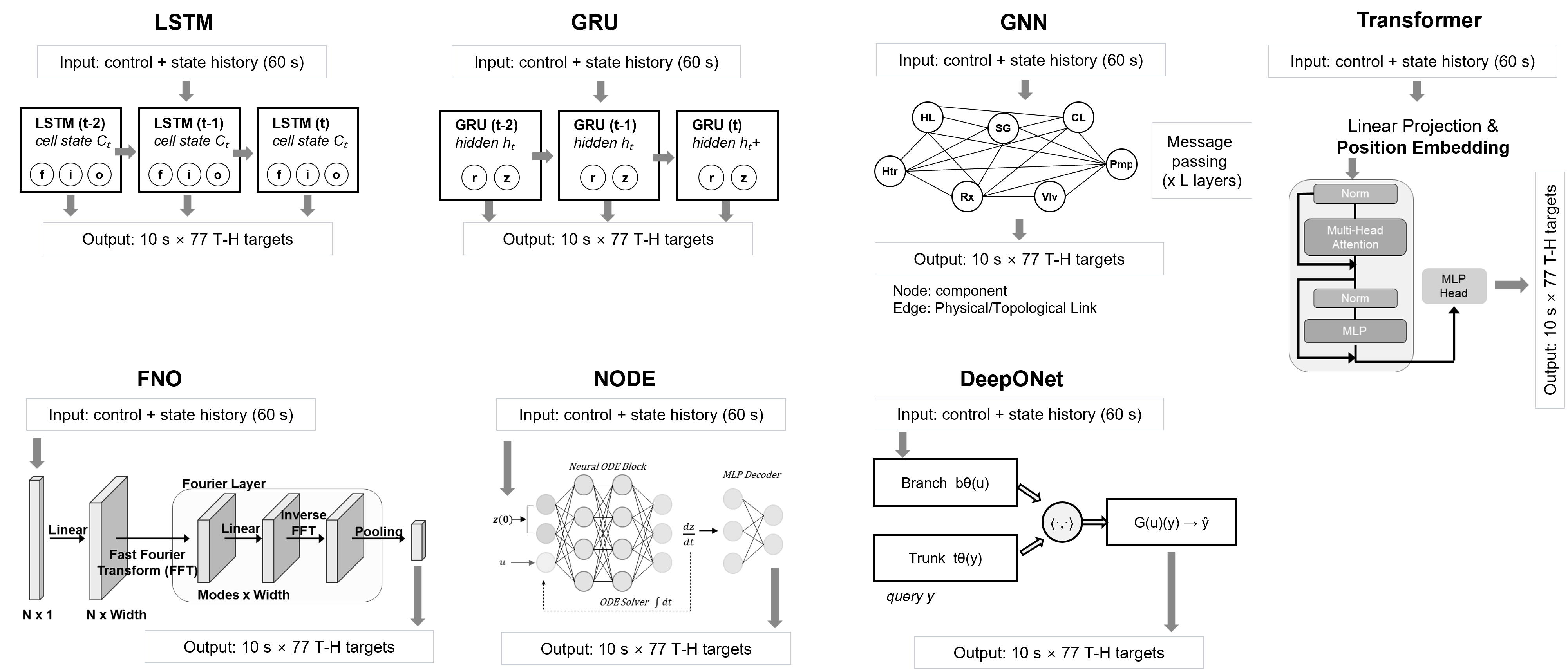}
  \caption{The seven candidate surrogate architectures screened offline.}
  \label{fig:pretrain_arch}
\end{figure}

As shown in Figure~\ref{fig:pretrain_arch}, the offline candidate pool contains seven model families: LSTM \citep{hochreiter1997lstm}, GRU \citep{cho2014gru}, Transformer \citep{vaswani2017attention}, neural ODE \citep{chen2018node}, graph neural network (GNN) \citep{battaglia2018graph,corso2024graph}, DeepONet \citep{lu2021deeponet}, and temporal Fourier neural operator (temporal-FNO) \citep{li2021fno}. These families were chosen because they impose different inductive biases on the same TH forecasting task. Recurrent networks provide compact temporal memory, Transformers represent long-range time coupling through attention, neural ODEs model continuous-depth latent dynamics, GNNs encode component-level coupling, and neural operators represent history-to-trajectory mappings.

Model selection uses blocked 3-fold cross-validation with a purged temporal gap of 70 windows to reduce leakage between adjacent time windows. The validation criterion is the mean absolute error over the 77 supervised targets. The model with the lowest cross-validation MAE is used only as the initial \emph{champion}; it is not assumed to remain optimal after the streaming regime changes. The architecture-search result and the pretraining-envelope trace check are reported in Section~\ref{sec:results_discussion} because they are performance results rather than method definitions.

\subsection{Streaming deployment and guarded continual adaptation}

After offline selection, each held-out stream split is processed sequentially from the same pretrained champion, and no promoted model state is carried from one split to the next. At each stream step, the current champion serves the 10-s forecast. When the corresponding measurements become available, the framework computes physical-unit error, warning/critical status, clipping fraction, top-error channels, and drift evidence. These monitoring signals can trigger candidate generation, but they do not directly replace the champion.

The online protocol follows the MLOps champion--challenger pattern for monitoring and updating evolving production ML systems~\citep{sculley2015mldebt,polyzotis2018datalifecycle}. The champion is the only model allowed to serve predictions. A challenger is a fine-tuned or retrained candidate evaluated against the incumbent champion using recent stream evidence and offline-plus-cumulative replay. The replay mixture, similar to Experience Replay and Maximally Interfered Retrieval~\citep{aljundi2019mir,buzzega2020der}, reduces overfitting to the most recent window and preserves earlier regimes, complementing the regularization-based remedy for catastrophic forgetting~\citep{kirkpatrick2017ewc}. Drift evidence is monitored online with an Adaptive Windowing detector~\citep{bifet2007adwin} that combines the strengths of CUSUM-style and binomial drift tests~\citep{page1954cusum,gama2004ddm}. Four controller actions are allowed: HOLD, ALARM, FINE\_TUNE, and RETRAIN. HOLD keeps the champion unchanged; ALARM records a warning without model replacement; FINE\_TUNE performs a bounded update around the current champion; and RETRAIN launches background challenger training. Promotion is allowed only when the candidate improves mean error without degrading exceedance, tail-risk, or clipping behavior; Table~\ref{tab:governance_rules} gives the full gate set. Rollback to the last stable champion is triggered by non-finite predictions or excessive clipping in the reported strict-gate suite, while a persistent-critical-error rollback trigger is defined in the broader safety envelope but disabled here. Algorithm~\ref{alg:guarded_loop} summarizes the resulting guarded streaming loop.

\begin{algorithm}[htbp]
  \caption{Guarded streaming adaptation loop}
  \label{alg:guarded_loop}
  \begin{algorithmic}[1]
    \State Select the initial champion $f_{\theta_0}$ by blocked cross-validation on the offline set.
    \State Initialize the stream buffer, replay buffer, stable-champion archive, cooldown counters, and audit log.
    \For{each evaluated stream window $t$}
    \State Serve $\widehat{\mathbf{Y}}_t$ with the current champion and apply physical clipping bounds.
    \State When measurements mature, compute channel-averaged MAE/RMSE, warning and critical status, clipping fraction, top-error targets, and drift evidence.
    \State Select HOLD, ALARM, FINE\_TUNE, or RETRAIN using deterministic rules and, when enabled, agent governance.
    \If{FINE\_TUNE is accepted by cooldown and data-availability checks}
    \State Perform bounded fine-tuning with replay mixing, weight blending, output-shift checking, and post-update validation.
    \EndIf
    \If{RETRAIN is accepted}
    \State Train background challengers from the offline replay buffer plus the cumulative stream buffer.
    \State Evaluate completed challengers against the incumbent champion on recent and replay evidence.
    \EndIf
    \State Promote only challengers that pass mean-error, exceedance, tail-risk, and clipping gates.
    \State Roll back to the last stable champion on non-finite predictions or excessive clipping (critical-only rollback disabled in the reported strict-gate suite).
    \State Record decisions, model states, agent calls, promotions, vetoes, and rollback events in the audit log.
    \EndFor
  \end{algorithmic}
\end{algorithm}

\subsection{Operating modes and agent governance}

The streaming comparison uses seven operating modes (Table~\ref{tab:operating_modes}): Static, Rule-H, Shadow, Single-H, Single-Full, MA-H, and MA-Full. They form an ablation ladder from no adaptation to role-separated governance. Static freezes the offline champion. Rule-H adds deterministic adaptation rules without agent queries. Shadow adds scheduled background refresh of the strongest pretrained families. Single-H and Single-Full test single-agent planning with selective and more frequent exposure, respectively. MA-H tests selective role-separated governance. In this paper, MA-Full denotes stepwise multi-agent governance: shadow refresh plus Monitor, Diagnosis, Adaptation, Safety-Auditor, and Orchestrator roles that review monitoring context, candidate priority, vetoes, and promotion decisions at every evaluated stream step.

\begin{table*}[htbp]
  \centering
  \caption{Operating modes used in the streaming comparison.}
  \label{tab:operating_modes}
  \footnotesize
  \setlength{\tabcolsep}{4.5pt}
  \renewcommand{\arraystretch}{1.12}
  \begin{tabular}{@{}p{0.2\textwidth}p{0.25\textwidth}p{0.45\textwidth}@{}}
    \toprule
    Mode & Agent                                                                                                                                                       & Core definition \\
    \midrule
    Static
         & None
         & Frozen pretrained champion; no adaptation or promotion.                                                                                                                       \\

    Rule-H
         & None
         & Rule-based guarded adaptation using deterministic gates only.                                                                                                                 \\

    Shadow
         & None
         & Periodic background refresh of the top-3 pretrained challenger families.                                                                                                      \\

    Single-H
         & Single agent, selective
         & Single-agent LLM planner queried at selected hierarchical decision points.                                                                                                    \\

    Single-Full
         & Single agent, frequent
         & Single-agent LLM planner queried at most streaming decisions.                                                                                                                 \\

    MA-H
         & Five agents, selective
         & Role-separated council with LLM planning queried at selected decision points.                                                                                                 \\

    MA-Full
         & Five agents, every step
         & Background shadow refresh with Monitor, Diagnosis, Adaptation, Safety-Auditor, and Orchestrator review at every evaluated stream step.                   \\
    \bottomrule
  \end{tabular}
\end{table*}

In MA-H and MA-Full, role separation makes decision support inspectable. The design follows reasoning-and-acting and multi-agent collaboration patterns in which specialized roles exchange structured observations and recommendations~\citep{wei2022cot,yao2023react,shinn2023reflexion,park2023generativeagents,hong2024metagpt,guo2024llmma}. The Monitor agent summarizes rolling error, warning streaks, clipping behavior, and dominant target channels. The Diagnosis agent maps those signatures to model-staleness or coupling-change hypotheses. The Adaptation agent proposes bounded fine-tuning, challenger refresh, or hold actions. The Safety-Auditor checks whether proposed updates violate hard gates or introduce tail-risk regression. The Orchestrator consolidates these inputs into the final recommendation. The council can recommend actions and vetoes, but deterministic validation, clipping, promotion, and rollback gates remain authoritative. The modes differ in how often these roles are applied: Single-H and MA-H apply them selectively after hierarchical triggers, Single-Full applies single-agent planning more frequently, and MA-Full applies the role-separated council at every evaluated stream step.

\subsection{Safety gates and evaluation protocol}

All adaptive modes share the same deterministic safety envelope, so the comparison cannot confuse agent recommendations with looser update rules. Table~\ref{tab:governance_rules} lists the hard safeguards: warning/critical thresholds computed after inverse transformation, cooldowns that prevent re-intervention before an action's effect is observable, prediction clipping against physically implausible outputs, bounded fine-tuning with weight blending and output-shift checks, a promotion gate requiring mean-error improvement without exceedance, tail-risk, or clipping regression, and rollback on non-finite prediction or excessive clipping. A persistent-critical-error rollback trigger is part of the full safety envelope but is disabled in the reported strict-gate suite.

\begin{table*}[htbp]
  \centering
  \caption{Deterministic safety envelope shared by all adaptive modes (one ``chunk'' = one streaming evaluation window).}
  \label{tab:governance_rules}
  \footnotesize
  \setlength{\tabcolsep}{6pt}
  \renewcommand{\arraystretch}{1.20}
  \begin{tabular}{@{}l p{0.72\textwidth}@{}}
    \toprule
    Component       & Condition / threshold                                                                                                                              \\
    \midrule
    \multicolumn{2}{@{}l}{\textit{Error state}}                                                                                                                          \\
    Warning         & MAE $\geq 5$                                                                                                                                       \\
    Critical        & MAE $\geq 8$                                                                                                                                       \\
    \addlinespace[2pt]
    \multicolumn{2}{@{}l}{\textit{Adaptation triggers}}                                                                                                                  \\
    Fine-tune       & warning streak $\geq 4$ chunks and buffer $\geq 16$ samples                                                                                        \\
    Retrain         & critical streak $\geq 2$ chunks, or warning streak $\geq 8$ with drift/staleness evidence                                                          \\
    \addlinespace[2pt]
    \multicolumn{2}{@{}l}{\textit{Action constraints}}                                                                                                                   \\
    Cooldown        & 4 / 6 / 12 chunks after fine-tune / retrain / failed fine-tune                                                                                     \\
    Weight blending & blend updated weights 0.35 toward pre-update model; reject if output-shift ratio $> 0.35$                                                          \\
    Output clipping & per-channel bounds from offline range (margin 0.25, $\pm 6\sigma$); auto-rollback if clip fraction $> 0.02$                                        \\
    \addlinespace[2pt]
    \multicolumn{2}{@{}l}{\textit{Validation}}                                                                                                                           \\
    Post-update     & 16 mixed recent/replay samples; reject if non-finite or post-update clip $> 0.005$                                                                 \\
    Challenger gate & online evaluation for 2 chunks against current champion                                                                                            \\
    \addlinespace[2pt]
    \multicolumn{2}{@{}l}{\textit{Promotion \& rollback}}                                                                                                                \\
    Promote         & mean-MAE gain $\geq 0.02$ AND no regression in exceedance, RMSE, p95, tail-mean, or clip fraction                                                  \\
    Rollback        & non-finite predictions or excessive clipping $\rightarrow$ last stable champion; critical-only rollback disabled in the reported strict-gate suite \\
    \bottomrule
  \end{tabular}
\end{table*}

For a mode $m$, the reported channel-averaged MAE is
\begin{equation}
  \mathrm{MAE}_{m}=\frac{1}{NHD}\sum_{t=1}^{N}\sum_{h=1}^{H}\sum_{d=1}^{D}\left|Y_{t,h,d}-\widehat{Y}^{(m)}_{t,h,d}\right|,
\end{equation}
where $N$ is the number of evaluated stream windows, $H=10$ is the forecast horizon, and $D=77$ is the number of supervised targets. The relative gain over Static is
\begin{equation}
  \mathrm{Gain}_{m}=100\left(\frac{\mathrm{MAE}_{\mathrm{Static}}-\mathrm{MAE}_{m}}{\mathrm{MAE}_{\mathrm{Static}}}\right).
\end{equation}
The final evaluation reports channel-averaged MAE, RMSE, warning-threshold exceedance, relative gain over Static, promotion behavior, and scenario-level prediction traces. These quantities are interpreted over six paired statistical units: two held-out stream splits times three seeds, with each split restarted from the same pretrained champion. To quantify uncertainty with this small effective sample size, paired mean differences in MAE and exceedance are summarized with non-parametric bootstrap confidence intervals using $10^4$ resamples at the 95\% confidence level~\citep{efron1979bootstrap}.

\section{Results and Discussion}
\label{sec:results_discussion}

\subsection{Pretrained surrogate model}

\begin{figure}[htbp]
  \centering
  \includegraphics[width=0.8\linewidth]{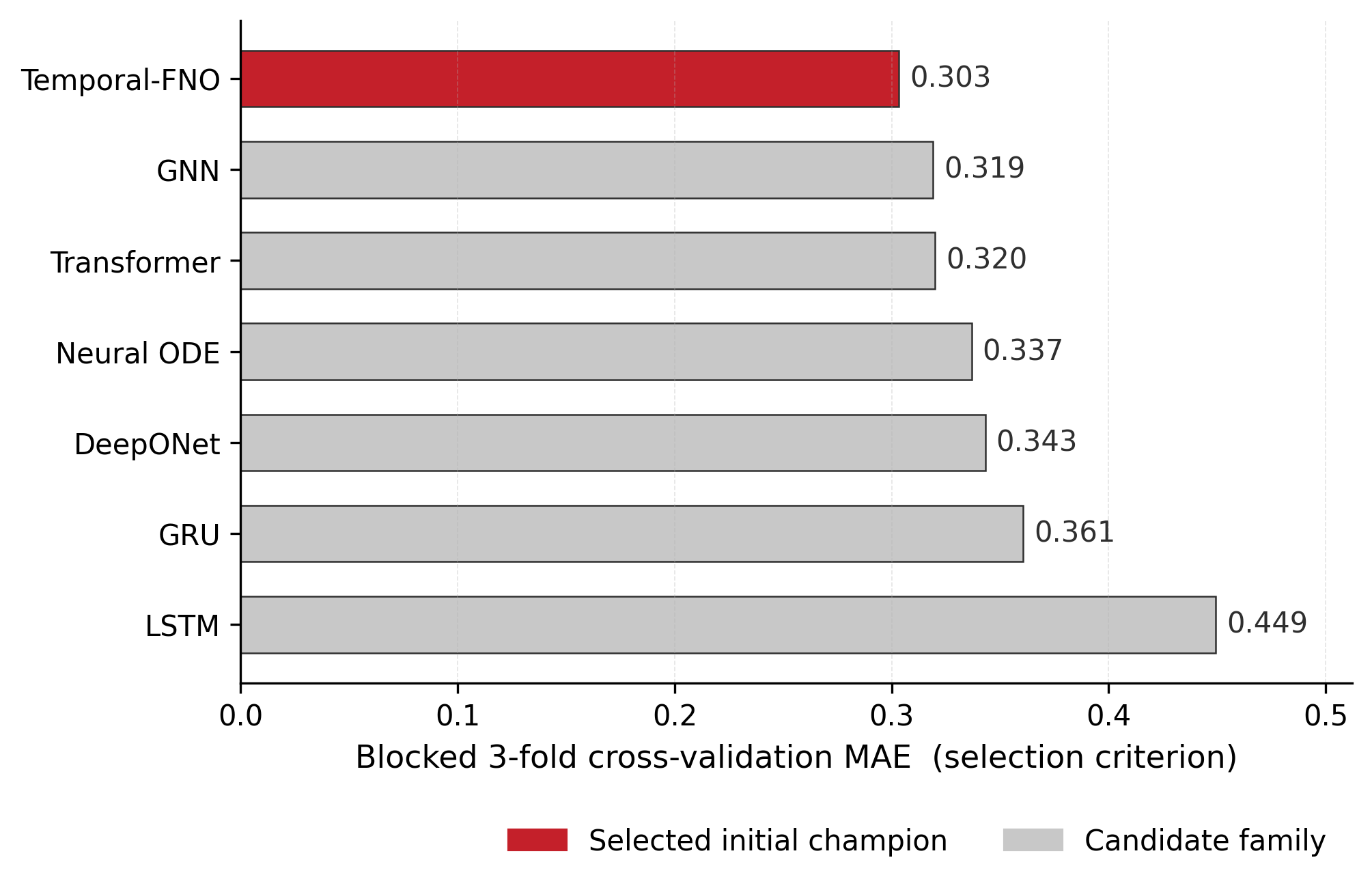}
  \caption{Offline champion selection by blocked 3-fold cross-validation. MAE is reported on the normalized target scale used for training and validation; temporal-FNO attains the lowest normalized CV MAE and is selected as the initial champion.}
  \label{fig:pretrain_search}
\end{figure}

Figure~\ref{fig:pretrain_search} summarizes the offline architecture search using the prespecified blocked 3-fold cross-validation criterion. The reported MAE values are computed on the normalized target scale used during training and validation. Temporal-FNO achieved the lowest normalized CV MAE of 0.303 and was therefore selected as the initial champion for all streaming comparisons. GNN and Transformer variants remained close enough to be retained as credible online challengers, but they were not used to define the initial static baseline.

\begin{figure}[htbp]
  \centering
  \includegraphics[width=\textwidth,height=0.90\textheight,keepaspectratio]{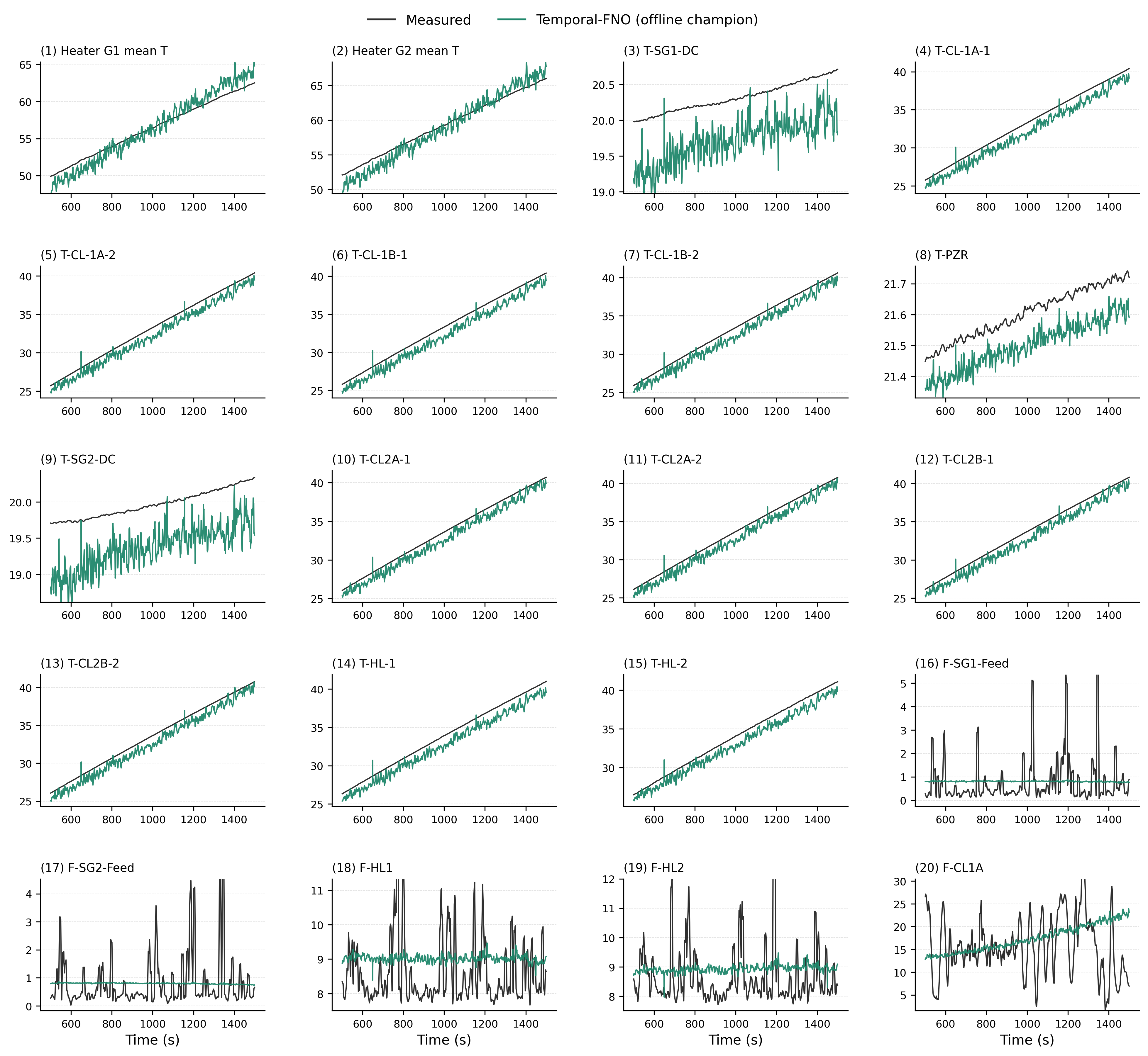}
  \caption{Temporal-FNO predictions on the pretraining-envelope control-rod transient.}
  \label{fig:pretrain_validation_controlrod}
\end{figure}

Figure~\ref{fig:pretrain_validation_controlrod} provides a trace-level check within the pretraining envelope. The temporal-FNO champion tracks the displayed subsets during the control-rod power-reduction transient with close agreement. This transient is part of the offline records, so close tracking indicates in-envelope capacity rather than out-of-envelope generalization. Once the heat-pipe configuration is introduced, the passive heat-removal pathway changes the dominant coupling structure, and the offline-selected champion becomes a candidate for adaptation.

\begin{figure}[htbp]
  \centering
  \includegraphics[width=\linewidth]{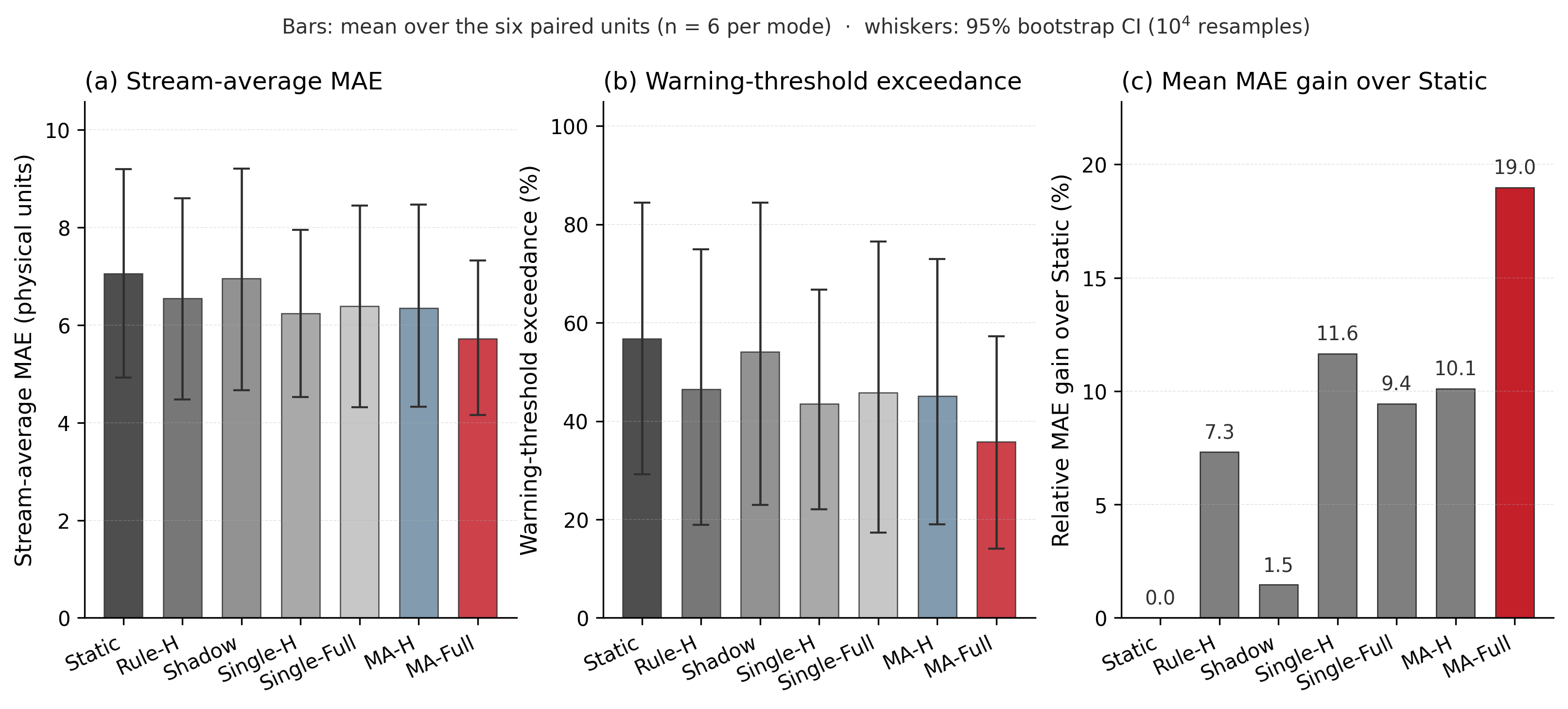}
  \caption{Multi-seed full-stream comparison across the seven modes: (a)~MAE and (b)~warning exceedance (bars: mean over six paired units; whiskers: 95\% bootstrap CI), and (c)~mean-MAE gain over Static.}
  \label{fig:accuracy}
\end{figure}

Table~\ref{tab:fullstep} and Figure~\ref{fig:accuracy} report the multi-seed full-stream comparison over two held-out stream splits. The Static mode keeps the offline temporal-FNO champion unchanged and obtains a channel-averaged MAE of $7.06 \pm 3.51$ with $56.8 \pm 45.4\%$ warning-threshold exceedance. This behavior is consistent with the condition-locking problem: a model that is accurate within the pretraining envelope can lose reliability when the facility enters a regime not represented during offline selection.

Rule-H reduces MAE to $6.54 \pm 3.30$, a 7.3\% gain over Static, without any agent query. In Table~\ref{tab:governance_rules}, the deterministic gates alone recover part of the lost performance, so Rule-H is a credible lower bound for adaptive deployment. Shadow, a harder non-agent baseline, improves only slightly ($6.96 \pm 3.63$), indicating that background challenger refresh by itself does not guarantee better deployment accuracy.

The lowest mean MAE and exceedance are obtained by MA-Full. MA-Full reduces MAE to $5.72 \pm 2.39$, improves mean MAE over Static by 19.0\%, and lowers warning exceedance to $35.8 \pm 31.5\%$. The paired six-unit analysis gives MA-Full mean differences of $-1.34$ in MAE and $-0.210$ in exceedance ratio relative to Static; 95\% non-parametric bootstrap intervals~\citep{efron1979bootstrap} for both differences exclude zero. Because intervals among adaptive modes overlap, this indicates that shadow learning combined with stepwise role-separated governance reduced mean error in the tested streams, while the comparison does not establish a definitive ranking among the adaptive modes. Single-H, Single-Full, and MA-H remain beneficial relative to Static but do not match MA-Full under the strict-gate suite.

\begin{table}[htbp]
  \centering
  \caption{Multi-seed full-stream comparison (mean $\pm$ s.d.\ over six paired units; Gain relative to Static). Exceedance is a fraction bounded in $[0,100]\%$; its standard deviation is a descriptive dispersion measure across the two heterogeneous splits, so the reported mean $\pm$ s.d.\ is not a symmetric confidence interval and can extend beyond the physical range.}
  \label{tab:fullstep}
  \footnotesize
  \setlength{\tabcolsep}{5.0pt}
  \begin{tabular}{lcccc}
    \toprule
    Mode        & MAE           & RMSE            & Exceed. (\%)  & Gain (\%) \\
    \midrule
    Static      & $7.06\pm3.51$ & $35.38\pm19.49$ & $56.8\pm45.4$ & 0.0       \\
    Rule-H      & $6.54\pm3.30$ & $34.60\pm19.48$ & $46.4\pm44.1$ & 7.3       \\
    Shadow      & $6.96\pm3.63$ & $35.19\pm19.69$ & $54.1\pm48.4$ & 1.5       \\
    Single-H    & $6.24\pm2.75$ & $34.09\pm18.49$ & $43.5\pm33.2$ & 11.6      \\
    Single-Full & $6.39\pm3.16$ & $34.57\pm19.12$ & $45.8\pm41.4$ & 9.4       \\
    MA-H        & $6.34\pm3.07$ & $34.39\pm19.04$ & $45.1\pm38.7$ & 10.1      \\
    MA-Full     & $5.72\pm2.39$ & $33.18\pm17.62$ & $35.8\pm31.5$ & 19.0      \\
    \bottomrule
  \end{tabular}
\end{table}

\subsection{Operational behavior of MA-Full}

\begin{figure}[htbp]
  \centering
  \includegraphics[width=\linewidth]{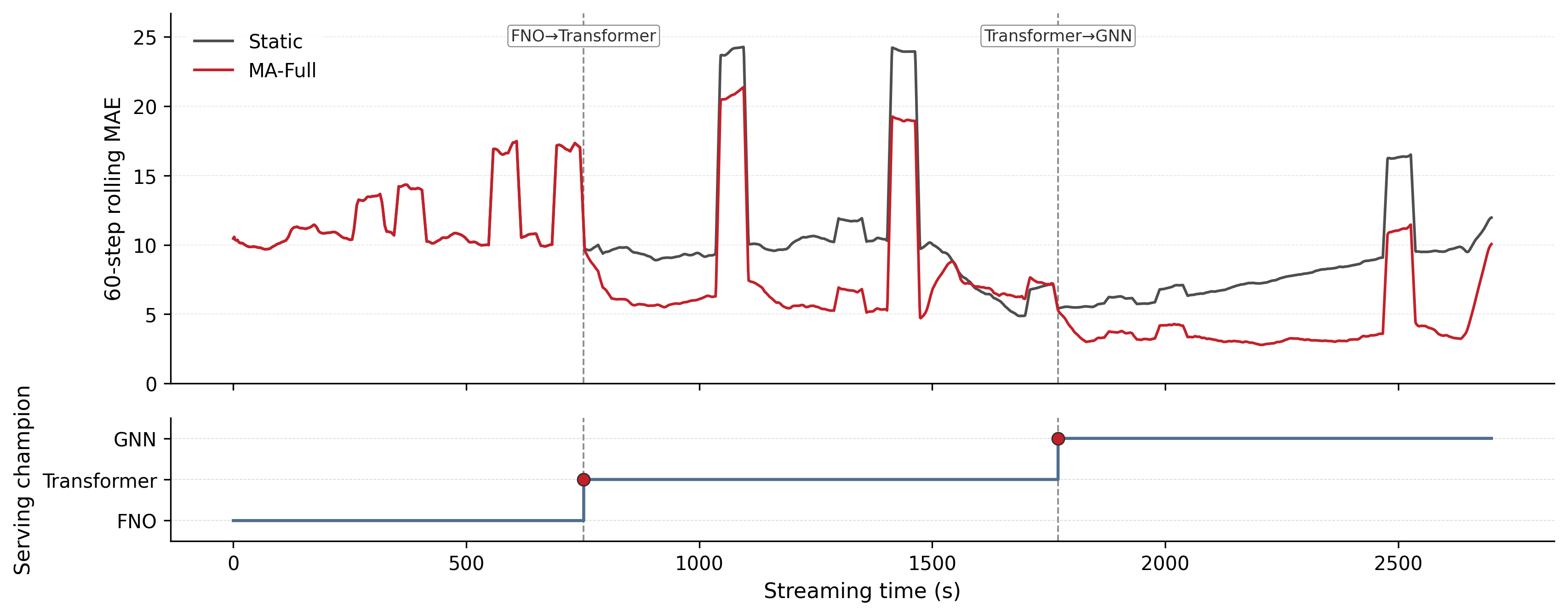}
  \caption{Rolling MAE (Static vs.\ MA-Full) and serving-champion state for the heat-pipe split; dashed lines mark validated transitions.}
  \label{fig:model_timeline_s1}
\end{figure}

\begin{figure}[htbp]
  \centering
  \includegraphics[width=\linewidth]{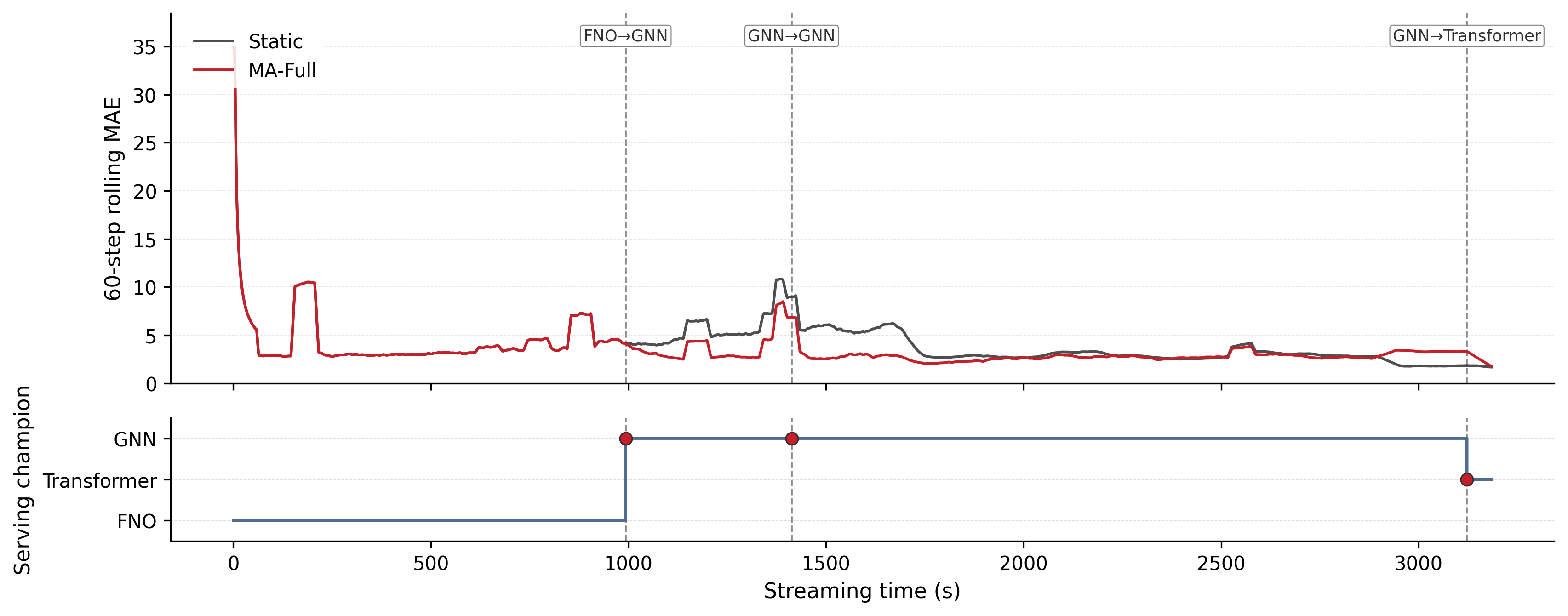}
  \caption{Rolling MAE and serving-champion state for the second held-out split (layout as in Figure~\ref{fig:model_timeline_s1}).}
  \label{fig:model_timeline_s2}
\end{figure}

The model-state timelines in Figures~\ref{fig:model_timeline_s1} and~\ref{fig:model_timeline_s2} align rolling-error behavior with champion transitions and pending challenger validation. MA-Full does not update whenever error rises: the serving model remains unchanged until a challenger passes validation, and the current champion continues serving when no validated challenger is available. The temporal alignment between some transitions and subsequent MAE recovery is consistent with stream-conditioned adaptation rather than a post-hoc average over unrelated periods.

\begin{figure}[t]
  \centering
  \includegraphics[width=\linewidth]{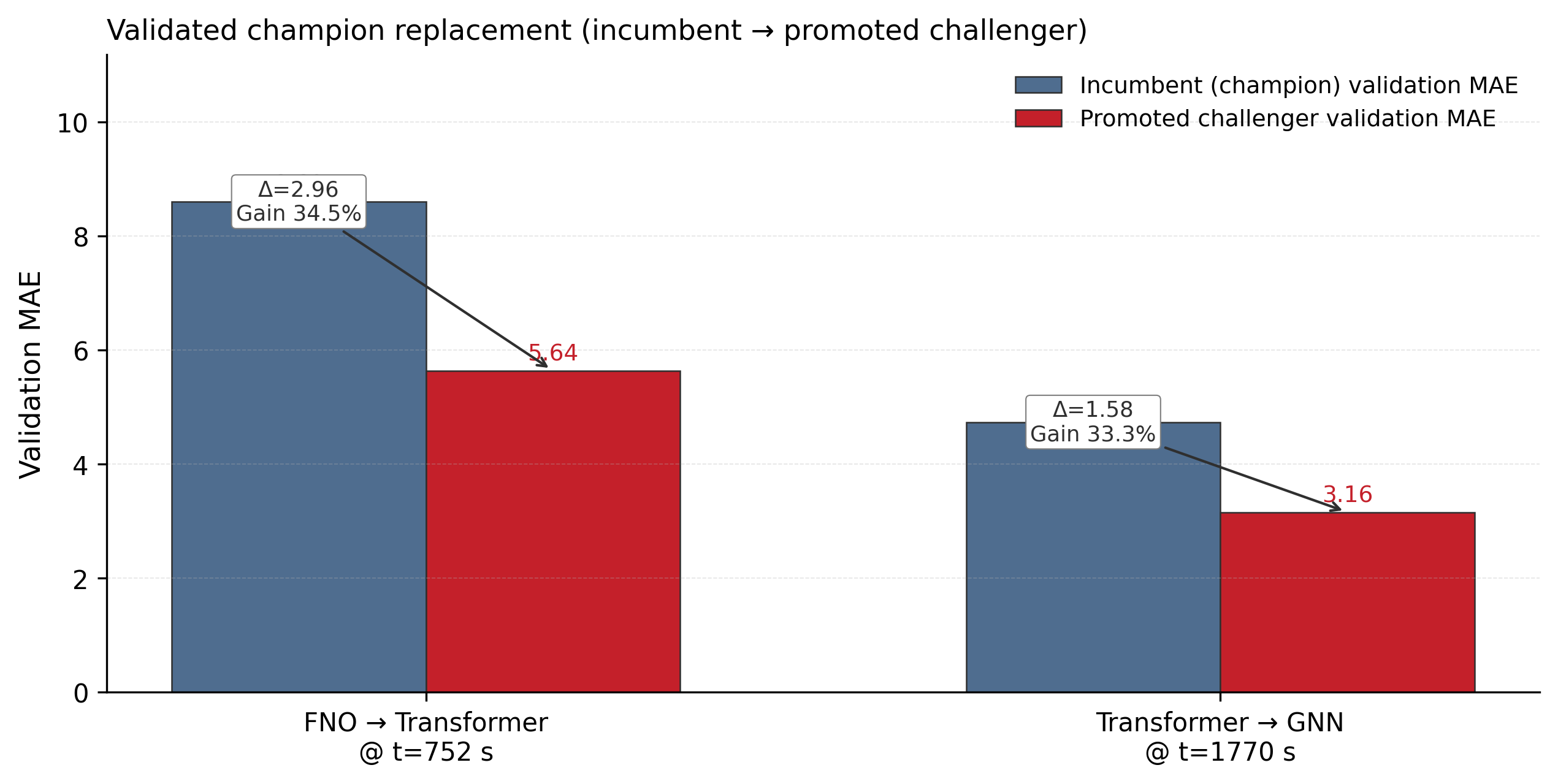}
  \caption{MA-Full validated champion replacements on the heat-pipe split: incumbent vs.\ promoted-challenger validation MAE at each transition.}
  \label{fig:fno_gnn_intervention}
\end{figure}

Table~\ref{tab:fno_gnn_intervention} and Figure~\ref{fig:fno_gnn_intervention} illustrate a central MA-Full behavior: instead of remaining fixed at the offline temporal-FNO champion, the served model moved through validated Transformer and GNN challengers when the streaming evidence supported replacement. This transition should not be interpreted as contradicting the offline architecture search. Temporal-FNO was the best model inside the pretraining envelope. The online Transformer and GNN challengers became useful only after the stream produced persistent evidence that the dominant error pattern involved delayed thermal memory, cross-variable hydraulic coupling, and flow-related channels.

In the representative heat-pipe split, validation promoted a compact Transformer over temporal-FNO and later a GNN over the Transformer (Figure~\ref{fig:fno_gnn_intervention}). Across all MA-Full runs, 13 validated champion transitions were logged---eight to GNN-family and five to Transformer-family challengers (Table~\ref{tab:fno_gnn_intervention}). MA-Full therefore did not choose a different family by preference; it used diagnosis-conditioned candidate priorities and gate-controlled validation to migrate the serving model only after stream-conditioned evidence supported the change.

\begin{table*}[htbp]
  \centering
  \caption{Champion surrogate model intervention summary of MA-Full mode.}
  \label{tab:fno_gnn_intervention}
  \small
  \setlength{\tabcolsep}{6pt}
  \renewcommand{\arraystretch}{1.15}
  \begin{tabular}{@{}p{0.25\textwidth}p{0.67\textwidth}@{}}
    \toprule
    Aspect & Key point                                                                                                                                                        \\
    \midrule
    Initial champion
           & Temporal-FNO was selected by blocked cross-validation as the strongest offline model (MAE 0.303).                                                                \\

    Drift evidence
           & The held-out streams produced persistent warning errors, mainly in flow and temperature channels.                                                                \\

    Candidate priority
           & Error signatures prioritized temporal-FNO, compact Transformer, dense-graph GNN, and graph-coupling GNN challengers.                                             \\

    Promotion evidence
           & In the representative heat-pipe split, FNO-to-Transformer validation reduced MAE from 8.60 to 5.64, followed by Transformer-to-GNN validation from 4.74 to 3.16. \\

    Repeated support
           & Across all MA-Full runs, 13 validated transitions were logged: eight to GNN-family challengers and five to Transformer-family challengers.                       \\

    Interpretation
           & The shifted streams favored attention-based temporal correction first and graph-based cross-variable coupling when hydraulic redistribution dominated.           \\
    \bottomrule
  \end{tabular}
\end{table*}

\begin{figure}[t]
  \centering
  \includegraphics[width=0.62\linewidth]{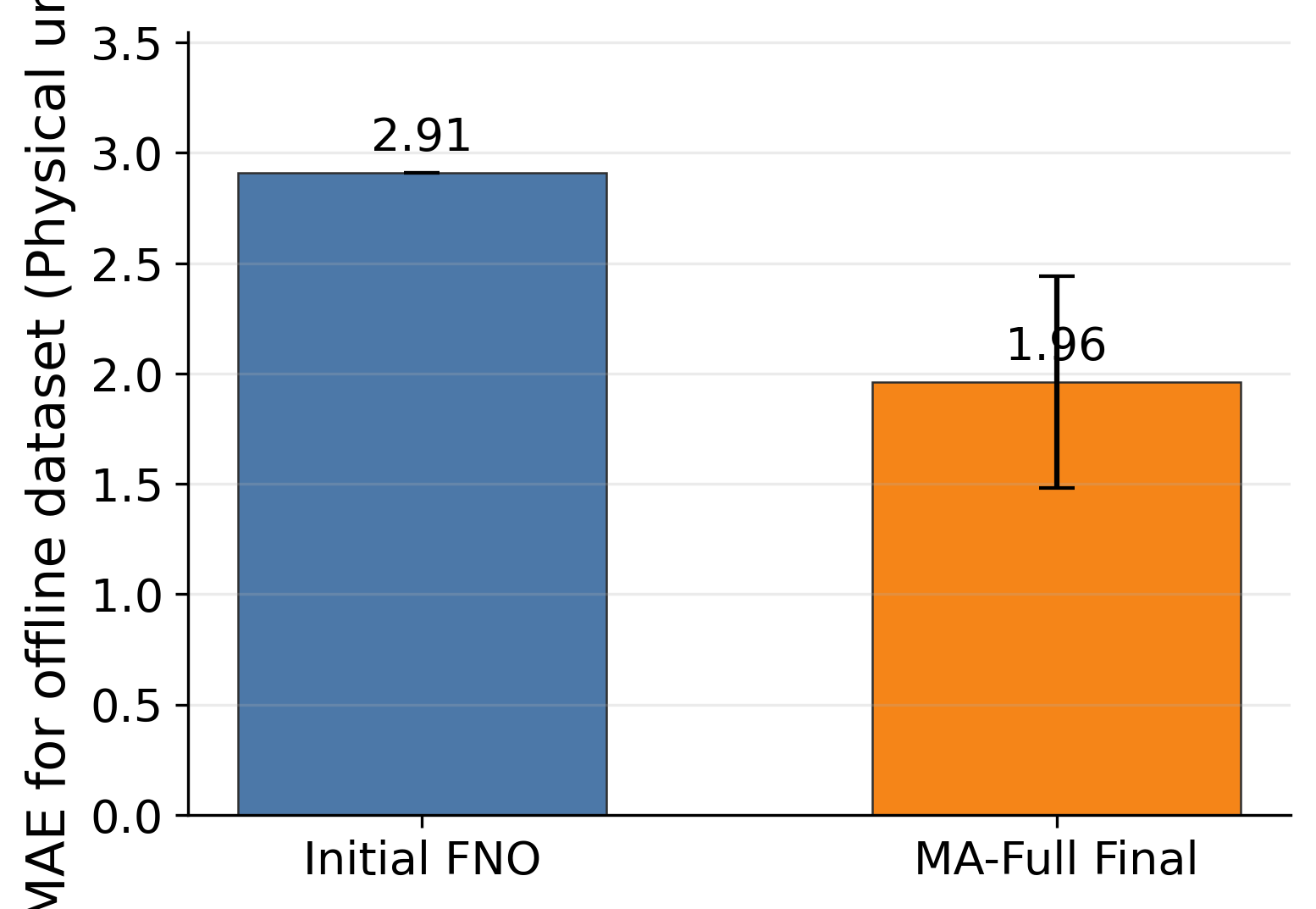}
  \caption{Offline-domain back-test after online adaptation. Channel-averaged MAE is evaluated after inverse transformation on all offline pretraining windows for the initial FNO champion and the final MA-Full champions.}
  \label{fig:offline_backtest}
\end{figure}

The same promotions were also checked against the original offline pretraining windows to test whether the online-adapted models simply overfit the held-out streams. Figure~\ref{fig:offline_backtest} compares the initial temporal-FNO champion with the final MA-Full champions after back-testing on all 8{,}343 offline training windows generated by the preprocessing pipeline from the offline records. The mean channel-averaged MAE decreased from 2.91 for the initial FNO to 1.96 for the final MA-Full champions, indicating that the replay-based challenger updates did not cause catastrophic forgetting of the pretraining envelope. This does not prove monotonic improvement for every target; some channels still degraded locally. However, it supports the intended continual-adaptation behavior: the served surrogate can improve under new stream evidence while retaining useful performance on the original operating domain.

\begin{figure}[htbp]
  \centering
  \includegraphics[width=\textwidth,height=0.86\textheight,keepaspectratio]{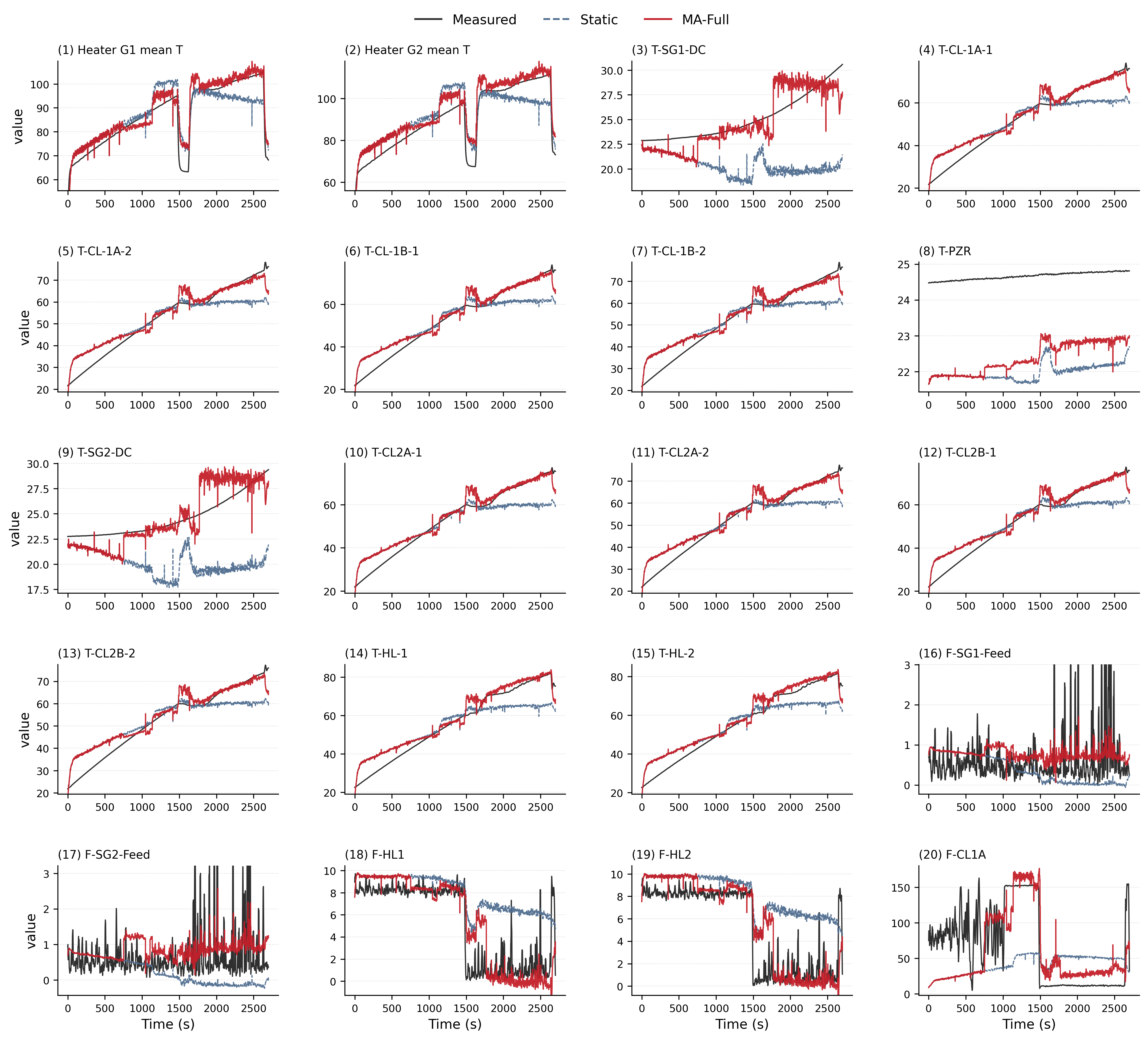}
  \caption{Variable-level prediction traces (measured, Static, MA-Full) for the heat-pipe split; each panel uses an independent 1st--99th-percentile vertical range.}
  \label{fig:scenario1_grid}
\end{figure}

\begin{figure}[htbp]
  \centering
  \includegraphics[width=\textwidth,height=0.86\textheight,keepaspectratio]{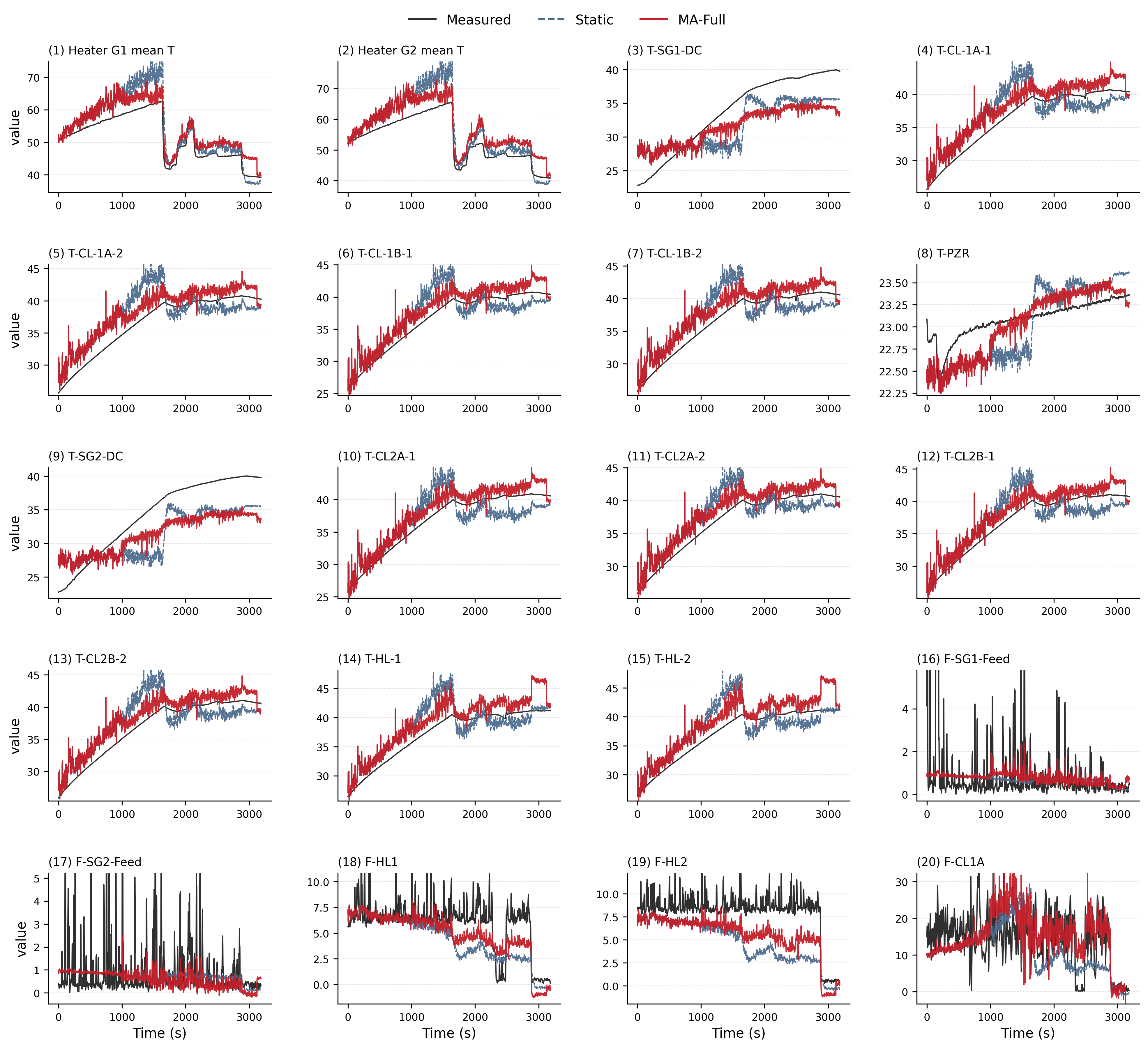}
  \caption{Variable-level prediction traces (measured, Static, MA-Full) for the second held-out split (layout as in Figure~\ref{fig:scenario1_grid}).}
  \label{fig:scenario2_grid}
\end{figure}

Figures~\ref{fig:scenario1_grid} and~\ref{fig:scenario2_grid} provide variable-level prediction traces for the two held-out stream splits using the same 20-panel state-channel convention as Figure~\ref{fig:pretrain_validation_controlrod}. In high-shift heat-pipe intervals, Static predictions gradually depart from several temperature and flow measurements, whereas MA-Full tends to recover closer tracking after validated adaptation. The improvement is not uniform across all targets or all times. Some intervals remain better served by the unchanged temporal-FNO model, and some channels retain large local errors. This mixed behavior is consistent with the full-stream metrics: governed continual adaptation reduces average deployment error, but it does not eliminate all channel-level extrapolation under unseen operating behavior.

\subsection{Multi-agent intervention behavior}

\begin{figure}[htbp]
  \centering
  \includegraphics[width=0.78\linewidth]{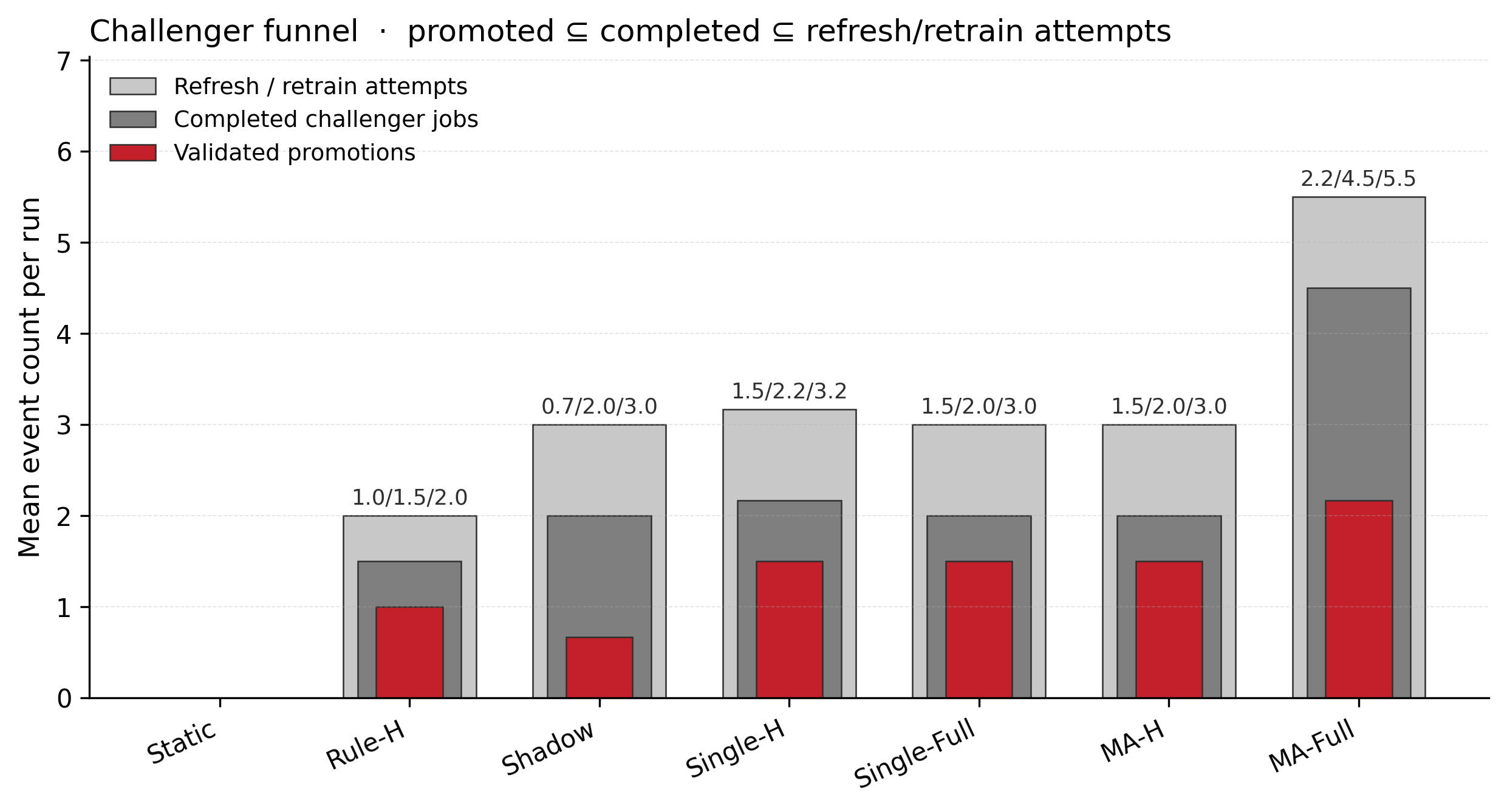}
  \caption{Challenger funnel across modes: validated promotions $\subseteq$ completed challenger jobs $\subseteq$ refresh/retrain attempts (mean per run; annotation reads promotions/completed/attempts).}
  \label{fig:actions_challengers}
\end{figure}

Adaptive modes repeatedly launch challenger training jobs, but only completed and validated challengers can be promoted. Figure~\ref{fig:actions_challengers} shows this as a funnel: refresh/retrain attempts, completed jobs, and validated promotions differ across modes, demonstrating that candidate exploration and model replacement are separate events. For TH deployment this separation matters because a challenger may be locally attractive during one stream segment but unsafe after the next regime fluctuation. Rollbacks and failed promotions are therefore expected safety outcomes; they are not implementation failures.

\subsection{Discussion}

The results support a cautious but useful conclusion. Agents add value mainly by governing which adaptation option is selected and when, while deterministic gates execute the change. Temporal-FNO is a strong offline model, yet the held-out streams still produce condition-locking behavior. Rule-H shows that deterministic thresholds, cooldowns, validation, and rollback already recover part of the lost performance. Shadow shows that background challenger learning alone is not enough. MA-Full performs best when these pieces are combined with role-separated review of monitoring context, candidate priority, vetoes, and promotions.

For safety-related TH monitoring, the proposed method is not an argument for autonomous replacement of the forecasting model whenever error rises. Serving-model replacement remains conservative, reversible, and evaluated on inverse-transformed measurements. Candidate models can be trained in the background, but deployment requires validation against mean error, exceedance, tail risk, and clipping behavior. In this sense the agent layer is a governance layer: it helps decide where adaptation effort should be spent and which candidate families deserve attention, while deterministic gates retain final authority.

The model-family transitions also clarify why a single offline architecture is unlikely to be sufficient. Temporal-FNO is well suited to the pretraining envelope, whereas the held-out heat-pipe and unseen-operation streams sometimes favor attention-based temporal correction or graph-based coupling. The remaining high-error intervals are dominated by flow-rate and hydraulically coupled channels, suggesting that future challengers should include explicit component connectivity, pump/valve-state constraints, pressure-drop structure, and conservative flow-balance penalties, consistent with physics-informed machine learning~\citep{karniadakis2021piml}. Long-horizon time-series architectures such as Informer or PatchTST~\citep{zhou2021informer,nie2023patchtst} could also be added without changing the surrounding governance layer.

Several limitations remain. The statistical evaluation uses two held-out stream splits and three seeds per split; the bootstrap intervals support the Static-versus-MA-Full comparison but do not establish universal superiority across all possible operating regimes. Broader claims require additional experimental campaigns, longer repeated streams, and ablations of clipping, bounded fine-tuning, tail-aware promotion, rollback, replay-buffer composition, and agent-role decomposition. Direct comparisons with continual-learning baselines such as experience replay, Maximally Interfered Retrieval~\citep{aljundi2019mir}, and EWC~\citep{kirkpatrick2017ewc} under the same safety envelope would further isolate the agent-governance contribution. Richer operator logs are also needed to connect agent diagnoses to known procedural events rather than only to error signatures.

\section{Conclusions}

This study developed and evaluated a guarded online-adaptation framework for state-feedback TH surrogate forecasting under operating-regime shift. A temporal-FNO model was selected as the initial champion by blocked 3-fold pretraining over seven candidate architectures. It was then deployed on two held-out stream splits under seven operating modes: Static, Rule-H, Shadow, Single-H, Single-Full, MA-H, and MA-Full.

The static champion achieved a channel-averaged MAE of $7.06 \pm 3.51$, illustrating that a strong offline surrogate can still degrade outside its pretraining envelope. Rule-H improved MAE to $6.54 \pm 3.30$ without agent queries, establishing deterministic guarded adaptation as a necessary baseline. Shadow refresh alone remained close to Static, whereas MA-Full achieved the best mean result: $5.72 \pm 2.39$ MAE and $35.8 \pm 31.5\%$ warning exceedance, corresponding to a 19.0\% mean MAE improvement over Static. Paired bootstrap intervals for MA-Full versus Static excluded zero in both MAE and exceedance, although the small number of paired units and overlapping intervals among adaptive modes limit the strength of broader claims.

The validated temporal-FNO-to-Transformer and Transformer-to-GNN promotions indicate that the served surrogate can evolve when candidate improvements pass strict mean-error, exceedance, tail-risk, and clipping gates. These results indicate that multi-agent governance can record candidate evaluation, veto, promotion, and rollback decisions while deterministic gates retain authority: agents diagnose error signatures, prioritize candidate families, and review risky promotions, but they do not replace validation. Future work should expand the stream library, include stronger physics-aware challengers, and compare the governance layer against standard continual-learning baselines under the same safety envelope.

\section*{Nomenclature}
\begin{tabular}{@{}ll}
  $t$                      & Streaming time index                                       \\
  $N$                      & Number of evaluated stream windows                         \\
  $H$                      & Forecast horizon; $H=10$ in this study                     \\
  $D$                      & Number of supervised target channels; $D=77$ in this study \\
  $\mathbf{X}_t$           & Input history window at time $t$                           \\
  $\mathbf{Y}_t$           & Measured future target trajectory                          \\
  $\widehat{\mathbf{Y}}_t$ & Predicted future target trajectory                         \\
  $\mathbf{u}$             & Control and power-related variables                        \\
  $\mathbf{z}$             & Measured thermal--hydraulic state variables                \\
  $f_{\theta}$             & Surrogate model with parameters $\theta$                   \\
  $\theta$                 & Trainable model parameters                                 \\
\end{tabular}

\section*{Abbreviations}
\begin{tabular}{@{}ll}
  AI          & Artificial Intelligence          \\
  TH          & Thermal--Hydraulic               \\
  LLM         & Large Language Model             \\
  MAE         & Mean Absolute Error              \\
  RMSE        & Root Mean Square Error           \\
  SBO         & Station Blackout                 \\
  CV          & Cross-Validation                 \\
  LSTM        & Long Short-Term Memory           \\
  GRU         & Gated Recurrent Unit             \\
  GNN         & Graph Neural Network             \\
  FNO         & Fourier Neural Operator          \\
  DeepONet    & Deep Operator Network            \\
  Rule-H      & Rule-based Hierarchical mode     \\
  Single-H    & Single-agent Hierarchical mode   \\
  Single-Full & Single-agent Full-exposure mode  \\
  MA-H        & Multi-Agent Hierarchical mode    \\
  MA-Full     & Multi-Agent Full-governance mode \\
\end{tabular}

\section*{Acknowledgements}
This work was supported by the Korea Institute of Energy Technology Evaluation and Planning(KETEP) grant funded by the Korea government(MOTIE)(No.RS-2024-00403194) and supported by the National Research Foundation of Korea(NRF) grant funded by the Korea government(MSIT) (No.RS-2026-25470750).

\section*{Data Availability}
The data presented in this research are available from the corresponding author upon reasonable request. All requests are subject to review prior to data release.

\section*{Conflicts of Interest}
The authors declare no conflicts of interest.

\section*{Declaration of generative AI}
During the preparation of this work, the authors used ChatGPT to assist with language editing. After using this tool, the authors reviewed and edited the content as needed and take full responsibility for the content of the publication.

\bibliographystyle{elsarticle-num}

\bibliography{references}

\end{document}